\documentclass[]{sjtu_sai}
\usepackage[utf8]{inputenc}
\usepackage{amsmath}
\usepackage{amsfonts}
\usepackage{nicefrac}
\usepackage{makecell}
\usepackage{colortbl}
\usepackage{pgfplots}
\usepackage{needspace}
\usepackage{wrapfig}
\pgfplotsset{compat=1.18}
\hypersetup{pdfauthor={Tianyi Yang; Haojun Yang; Jipeng Xiao; Xuesong Chen; Shaoshuai Shi; Li Jiang; Zhipeng Zhang},pdftitle={Drive3R: Block-wise Streaming Geometric Foundation Models for Autonomous Driving}}
\title{Drive3R: Block-wise Streaming Geometric Foundation Models for Autonomous Driving}
\renewcommand{\authorfont}{\fontsize{12}{14}\selectfont}
\renewcommand{\affiliationfont}{\fontsize{9.5}{11.5}\selectfont}
\renewcommand{\affiliation}[2][]{\addtolist[#1]{#2}{\affiliationlist}{\affiliationformat}{\par}}
\renewcommand{\authorlist}{%
  \authorformat[1,2]{Tianyi Yang}\quad
  \authorformat[1]{Haojun Yang}\quad
  \authorformat[1]{Jipeng Xiao}\quad
  \authorformat[2]{Xuesong Chen}\par
  \authorformat[2]{Shaoshuai Shi}\quad
  \authorformat[3]{Li Jiang}\quad
  \authorformat[1,*]{Zhipeng Zhang}}
\affiliation[1]{AutoLab, School of Artificial Intelligence, Shanghai Jiao Tong University}
\affiliation[2]{Voyager Research, Didi Chuxing}
\affiliation[3]{The Chinese University of Hong Kong, Shenzhen}
\contribution[*]{Corresponding author.}
\renewcommand{\checkdatalist}{\relax}
\patchcmd{\mymaketitle}{\vskip 6mm}{\vskip 4mm}{}{\PackageError{Drive3R}{Author spacing patch failed}{}}
\patchcmd{\mymaketitle}{\vskip 3mm}{\vskip 1.5mm}{}{\PackageError{Drive3R}{Affiliation spacing patch failed}{}}
\patchcmd{\mymaketitle}{\vskip 3mm}{\vskip 1.5mm}{}{\PackageError{Drive3R}{Correspondence spacing patch failed}{}}
\patchcmd{\mymaketitle}{\vskip 6mm}{\vskip 4mm}{}{\PackageError{Drive3R}{Abstract spacing patch failed}{}}

\fancypagestyle{firststyle}{%
  \fancyhf{}%
  \fancyhead[L]{%
    \raisebox{-0.5\height}{\includegraphics[viewport=102 142 1169 1171,clip,height=12mm,keepaspectratio]{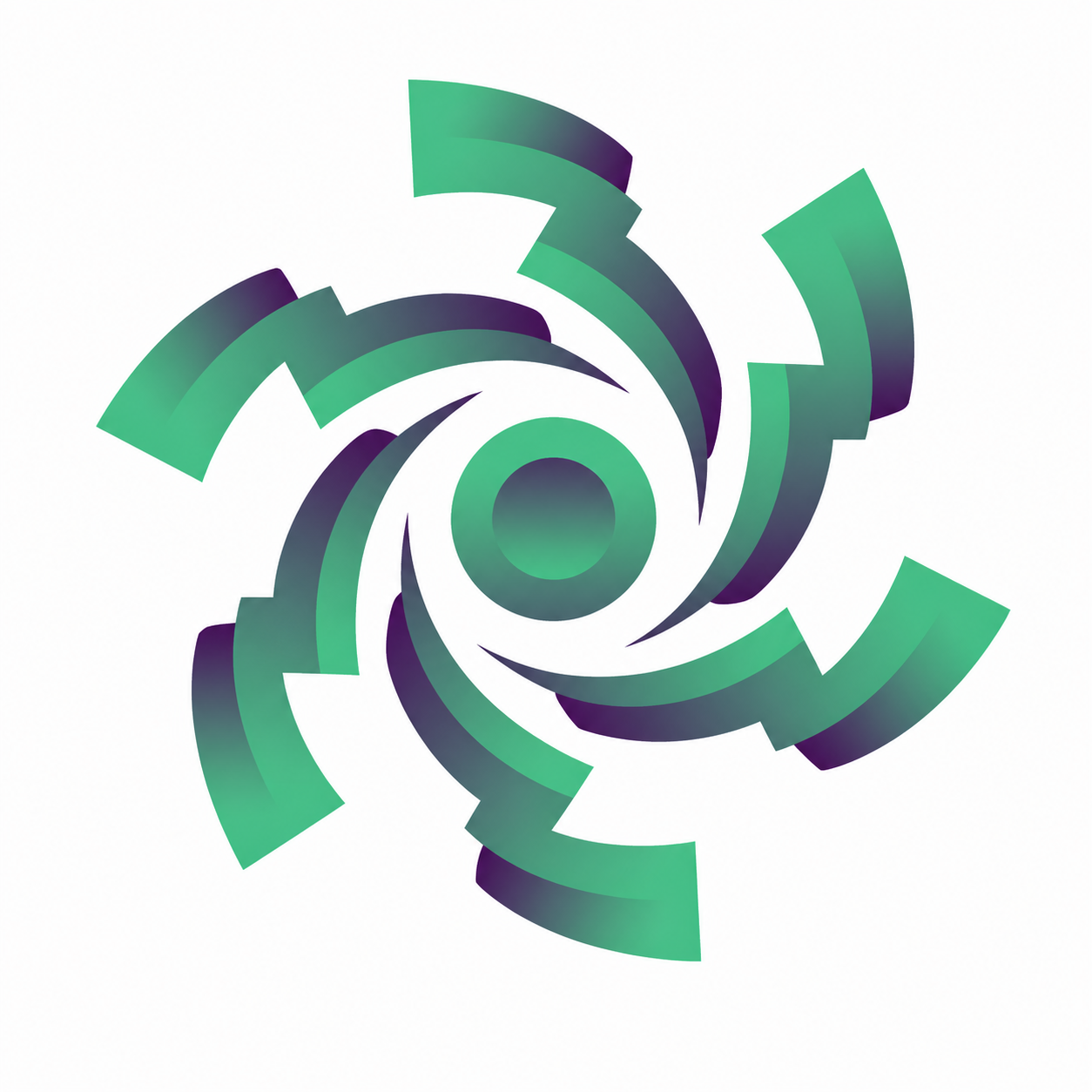}}\hspace{3.5mm}%
    \raisebox{-0.5\height}{\includegraphics[viewport=4 4 398 398,clip,height=12mm,keepaspectratio]{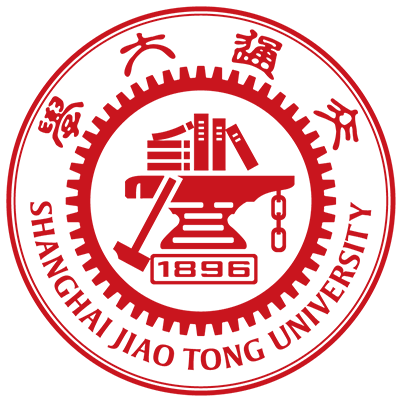}}\hspace{3.5mm}%
    \raisebox{-0.5\height}{\includegraphics[height=12mm,keepaspectratio]{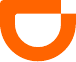}}\hspace{3.5mm}%
    \raisebox{-0.5\height}{\includegraphics[viewport=0 0 128 100,clip,height=12mm,keepaspectratio]{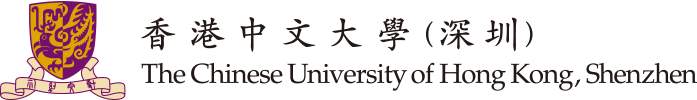}}%
  }%
  \fancyfoot[C]{\thepage}%
}

\abstract{High-fidelity closed-loop simulation for autonomous driving increasingly relies on dense multi-view 3D reconstruction. While recent feed-forward geometric foundation models have revolutionized per-scene optimization, applying them to continuous, long-horizon driving sequences is computationally intractable due to GPU memory limits and the quadratic complexity of global attention. Existing scaling strategies are fundamentally limited, as causal streaming suffers from compounding scale drift and trajectory deviation, while rigid "chunk-then-align" heuristics fragment global context and produce trajectory discontinuities. To address these limitations, we introduce Drive3R, a block-wise streaming geometric foundation model. Its block-wise causal attention resolves the scalability bottleneck of fully bidirectional attention while avoiding the loss of local future context in purely causal streaming. Its confidence-aware re-anchoring further mitigates long-range drift by detecting unreliable streaming states, re-initializing the KV cache, and stitching refreshed segments through $\mathrm{SE}(3)$ alignment. Extensive experiments demonstrate that Drive3R establishes a new state-of-the-art on nuScenes, KITTI, and Waymo. Specifically, on nuScenes, it improves over the recent DVGT-2 baseline from 0.1806 to 0.1081 in AbsRel, from 0.6750 to 0.5344 in point-cloud Accuracy, and from 1.5812 to 0.9092 in ATE. \newline Project page: \url{https://silence-ai423.github.io/Drive3R/}.}

\begin{document}
\maketitle

\begin{figure}[!t]
  \centering
  \includegraphics[width=\linewidth]{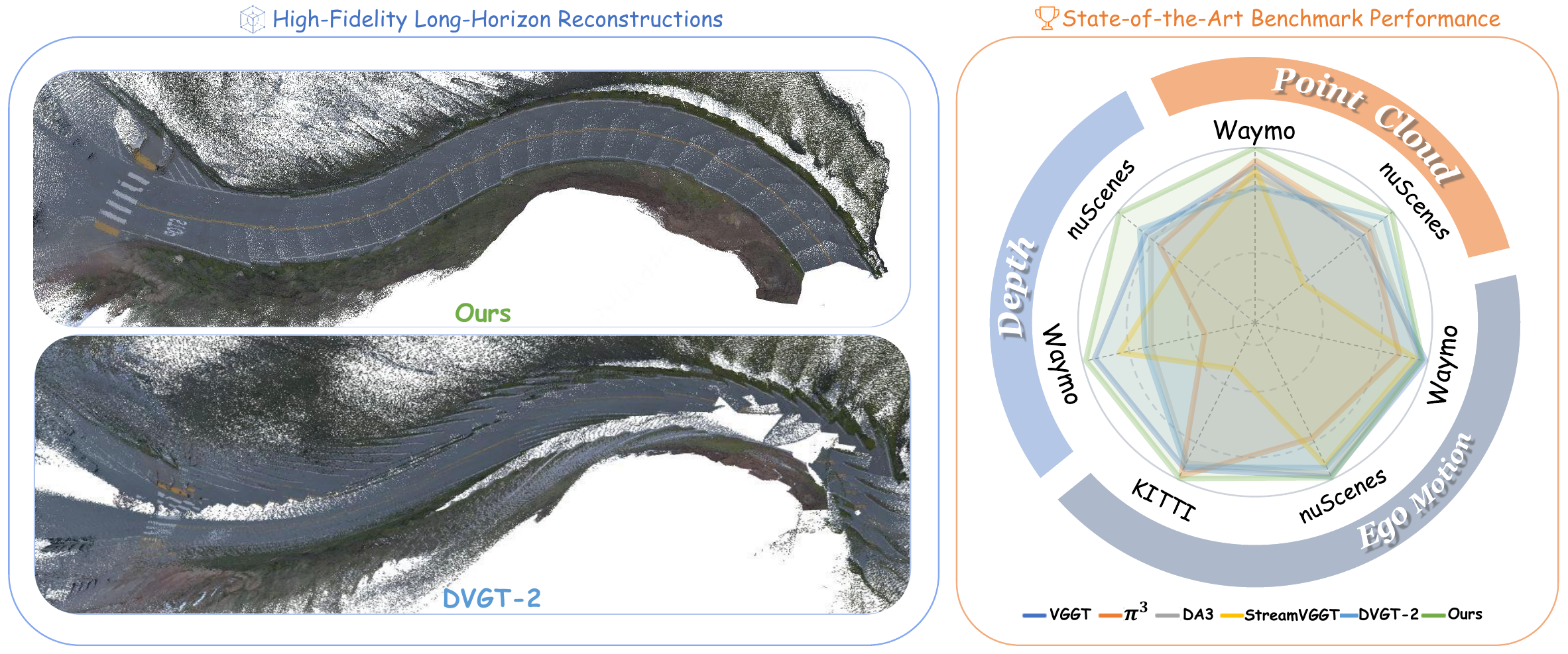}
\caption{\textbf{Teaser of Drive3R.}
Compared with the per-frame streaming baseline DVGT-2 \cite{zuo2026dvgt}, Drive3R reconstructs cleaner and more coherent long-horizon driving scenes with better preserved road geometry. The radar chart further shows state-of-the-art performance across depth, point-cloud, and ego-motion metrics on nuScenes \cite{caesar2020nuscenes}, Waymo \cite{sun2020scalability}, and KITTI \cite{geiger_vision_2013}.}

  \label{fig:teaser}
\end{figure}

\section{Introduction}

The recent leap in end-to-end autonomous driving~\cite{hu2023planning,guo2025end,zhou2025autovla, zhang2026efficient, zhang2026onedrive} demands high-fidelity, closed-loop simulation for scalable training and evaluation. As traditional graphics-based rendering struggles to synthesize photorealistic sensor data, modern reconstruction techniques like 3D Gaussian Splatting (GS) have emerged as the de facto standard for building true-to-life digital twins \cite{kerbl20233d, zhou2024drivinggaussian, yan2024street}. While early pipelines rely on LiDAR for geometric initialization, its inherent sparsity severely limits the density and completeness of the reconstructed environments \cite{tian2023occ3d, tonderski2024neurad}. Consequently, extracting dense and accurate 3D geometry purely from multi-view 2D images is no longer just an alternative, but a critical prerequisite for next-generation driving neural simulators \cite{yu2026recondrive, shi2025unisplat}.

\begin{figure}[!b]
  \centering
  \includegraphics[width=\linewidth]{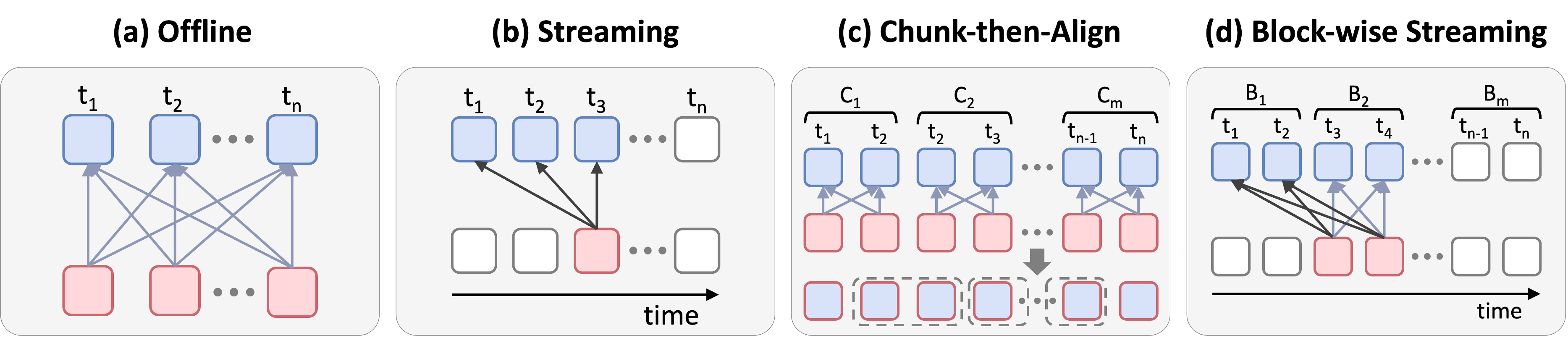}
  \caption{Paradigm comparison between previous sequence processing strategies (offline, streaming, chunk-then-align) and our proposed block-wise streaming approach, Drive3R.}
  \label{fig:paradigm_comparison}
\end{figure}

To achieve this dense geometry recovery, traditional multi-view methods typically rely on \textit{per-scene} global iterative optimization \cite{schonberger2016structure, yao2018mvsnet, mildenhall2021nerf}, which is computationally prohibitive and scales poorly to large-scale applications. Recently, geometric foundation models, typified by VGGT \cite{wang2025vggt,shen2025fastvggt}, have revolutionized this process by introducing a \textit{feed-forward} prediction paradigm \cite{wang2024dust3r, hong2023lrm}. By directly outputting dense point clouds through a single forward pass, these models offer a leap in computational efficiency. However, autonomous driving inherently involves continuous, long-horizon video streams, where a single driving log can easily span thousands of frames. As illustrated in Fig.~\ref{fig:paradigm_comparison}a, when confronted with such extensive sequences, existing fully bidirectional feed-forward models hit a hard wall due to the severe constraints of GPU memory limits and the quadratic complexity of global attention mechanisms~\cite{shen2025fastvggt}. Consequently, processing an entire long sequence via a single forward inference pass becomes computationally intractable.

To bypass this computational bottleneck, recent efforts to scale feed-forward models for long sequences have predominantly diverged into two distinct routes. The first route enforces a strict causal structure for per-frame streaming prediction (Fig.~\ref{fig:paradigm_comparison}b) \cite{wang2025continuous, chen2025ttt3r, zhuo2025streaming, yuan2026infinitevggt}, while the second relies on a ``chunk-then-align'' (Fig.~\ref{fig:paradigm_comparison}c) \cite{deng2025vggt, lin2025depth}. Yet, when deployed in complex, long-range autonomous driving scenarios, both paradigms reveal fundamental flaws. By entirely discarding bidirectional context, causal streaming suffers from the continuous accumulation of minor historical errors. Over extended sequences, these errors compound, inevitably triggering severe scale drift and trajectory deviation. Conversely, while the ``chunk-then-align'' approach preserves bidirectional attention within isolated chunks, its rigid and coarse-grained mechanism struggles to adapt to highly dynamic environments, as predefined chunks often blindly sever continuous motions. Furthermore, the absence of coherent feature propagation between chunks forces a reliance on rigid post-processing alignment in overlapping regions, frequently producing jarring trajectory discontinuities at chunk boundaries.

In response to these fundamental limitations, we present Drive3R, a block-wise streaming geometric foundation model explicitly tailored for the long-horizon demands of autonomous (Fig.~\ref{fig:paradigm_comparison}d). Its architecture is driven by two synergistic innovations, specifically \textit{block-wise causal attention} and \textit{confidence-aware re-anchoring}. First, to combat the scale drift inherent in pure streaming, we introduce block-wise causal attention. By allowing tokens within each block to attend bidirectionally, this mechanism establishes a robust local geometric context, effectively halting the monotonic error amplification along the temporal axis. Concurrently, strict causality is enforced across blocks. Unlike the isolated ``chunk-then-align'' approach that fragments global context, this design leverages cross-block KV cache reuse to continuously propagate historical geometric features, thereby maintaining long-range consistency while guaranteeing a highly scalable streaming inference complexity. Second, to address the rigidity of fixed-length chunking, we propose confidence-aware re-anchoring. Equipped with an additional lightweight confidence head, our Drive3R dynamically predicts the accumulated drift relative to the initial KV cache state with minimal computational overhead. During online inference, if this confidence score drops below a critical threshold, indicating potential trajectory degradation, the model proactively discards the stale KV cache, backtracks a few frames to re-initialize a fresh block, and seamlessly stitches the new segment into the global trajectory via $\mathrm{SE}(3)$ alignment on overlapping frames. Ultimately, this adaptive mechanism ensures continuous, discontinuity-free geometry recovery even in highly dynamic environments.

By synergizing these two designs, Drive3R elegantly unifies intra-block bidirectional precision, cross-block streaming throughput, and drift-aware long-sequence stability within a single network. Extensive experiments validate that our approach establishes a new state-of-the-art for dense geometric recovery. Notably, Drive3R currently stands as the strongest performing algorithm on the challenging nuScenes dataset \cite{caesar2020nuscenes}. It significantly surpasses both the previously dominant offline baseline (DVGT) \cite{zuo2025dvgt} and its per-frame streaming counterpart (DVGT-2) \cite{zuo2026dvgt}, improving AbsRel from 0.1397/0.1806 to 0.1081, point-cloud Accuracy from 0.6158/0.6750 to 0.5344, and ATE from 1.1012/1.5812 to 0.9092.

In summary, our main contributions are as follows: $\spadesuit$ \textbf{A novel block-wise streaming paradigm.} We introduce Drive3R, a geometric foundation model that resolves the fundamental conflict between sequence length and geometric consistency. By leveraging block-wise causal attention, it establishes robust local contexts bidirectionally while continuously propagating historical features via cross-block KV cache reuse, guaranteeing a highly scalable inference complexity. $\spadesuit$ \textbf{Dynamic re-anchoring.} To combat scale drift and the rigidity of fixed-length chunking, we propose an adaptive re-anchoring mechanism. This module predicts temporal drift online and proactively re-initializes the KV cache upon detecting instability, seamlessly stitching new segments to ensure discontinuity-free long-range trajectory recovery. $\spadesuit$ \textbf{Unprecedented state-of-the-art performance.} Drive3R currently stands as the strongest performing algorithm on multiple challenging autonomous driving datasets, including nuScenes \cite{caesar2020nuscenes}, KITTI \cite{geiger_vision_2013}, and Waymo \cite{sun2020scalability}. It comprehensively surpasses previously dominant offline, per-frame streaming and chunk-then-align baselines across all core geometric metrics, including ego-pose estimation, dense depth, and point cloud reconstruction.

\section{Related Work}
\textbf{Feed-Forward 3D Reconstruction Foundation Models.} While traditional optimization-based methods achieve high-fidelity geometry recovery \cite{schonberger2016structure}, their computationally prohibitive iterative nature makes them difficult to scale for large-scale applications. To circumvent these limitations, a class of feed-forward 3D reconstruction Transformers has rapidly emerged, pioneered by DUSt3R \cite{wang2024dust3r}. In a single forward pass, they directly regress pixel-aligned pointmaps, depths, and camera poses from images, effectively bypassing the arduous iterative optimization process. Subsequent works have extended this from pair-based to joint multi-view inference \cite{tang2025mv,zhang2025review}. Recently, VGGT \cite{wang2025vggt,shen2025fastvggt} and $\pi^3$ \cite{wang2025pi} further scaled this paradigm to large-scale Transformer backbones, capable of reconstructing scenes comprising dozens of frames using a single global bidirectional attention, establishing themselves as the state-of-the-art general geometric foundation models. In autonomous driving scenarios, DVGT \cite{zuo2025dvgt} adapts this paradigm to surround-view multi-camera inputs, directly outputting metric-scale dense geometry and ego-vehicle poses, which can serve directly as a geometric foundation for downstream scene reconstruction. However, these models architecturally rely on global bidirectional attention, requiring the entire sequence to be processed in a single forward pass. The $\mathcal{O}(N^2)$ complexity of the attention mechanism \cite{shen2025fastvggt} makes them inherently incompatible with the continuous, long-sequence inference demanded by autonomous driving. This is precisely the bottleneck our paper aims to overcome while preserving the benefits of bidirectional reasoning.

\textbf{Long-Sequence Geometric Foundation Models.} To adapt feed-forward geometric reconstruction to long-duration scenarios, recent works mainly follow two trajectories. The first introduces a strict causal structure for per-frame streaming prediction. Specifically, CUT3R \cite{wang2025continuous} maintains an implicit scene state at each step, StreamVGGT \cite{zhuo2025streaming} transforms VGGT \cite{wang2025vggt} into a KV-cached causal Transformer, and DVGT-2 \cite{zuo2026dvgt} adopts a similar sliding-window causal structure. While incremental inference avoids repeated full-sequence reconstruction, these models completely sacrifice bidirectional context. By relying solely on historical priors without future lookahead, minute per-frame errors accumulate monotonically, inevitably leading to scale drift and trajectory deviation. The second trajectory adopts a ``chunk-then-align'' strategy, as seen in VGGT-Long \cite{deng2025vggt} and DA3-Stream \cite{lin2025depth,deng2025vggt}. These pipelines partition sequences into overlapping fixed-length chunks, perform independent offline bidirectional reconstruction on each, and subsequently align them via rigid $\mathrm{Sim}(3)$ transformations. Although this preserves local bidirectional attention, the predefined chunking often severs continuous motions in highly dynamic environments, and the absence of coherent feature propagation across chunks forces a reliance on post-processing, frequently causing trajectory discontinuities at boundaries. In contrast, Drive3R resolves these conflicting paradigms. By leveraging block-wise causal attention, it achieves highly scalable streaming without sacrificing local bidirectional reasoning. Coupled with a confidence-aware dynamic re-anchoring mechanism to correct accumulated drift online, Drive3R ensures robust, discontinuity-free long-range trajectory recovery.

\section{Method}
\subsection{Problem Formulation and Overview}
\label{sec:problem_formulation}

Given a long-horizon surround-view image stream in autonomous driving, we denote the input as
\begin{equation}
    \mathcal{I}_{1:T}=\{I_{t,v}\mid t=1,\ldots,T,\ v=1,\ldots,V\},
\end{equation}
where \(T\) is the sequence length and \(V\) is the number of surround-view cameras. For each image \(I_{t,v}\), the model is provided with calibrated camera intrinsics \(K_{t,v}\) and the camera-to-ego extrinsics \(T^{\mathrm{ego}}_{t,v}\). Our goal is to online predict the ego pose trajectory \(\{E_t\}_{t=1}^{T}\) and the dense 3D point clouds \(\{P_{t,v}\}_{t=1,v=1}^{T,V}\) aligned with the multi-camera observations.

Compared with generic multi-view reconstruction, geometric modeling for autonomous driving has more specific task constraints and stronger available priors. First, the input usually comes from continuous driving logs, whose sequence length is far beyond the short clips commonly considered in generic reconstruction. Therefore, the model must scale to long-horizon streaming inference. Second, downstream simulation rely on geometry in metric space, requiring the model to preserve absolute scale consistency rather than only recover relative structure. Third, autonomous driving platforms provide stable and calibrated camera intrinsics and extrinsics. Unlike generic scenes where camera parameters are unknown or need to be estimated, these calibration parameters can be explicitly injected into the model as reliable geometric priors, helping the network reason about spatial relationships among different camera observations.

\subsection{Overall Architecture}
\label{sec:overall_architecture}

\begin{figure}[!b]
    \centering
    \includegraphics[width=\textwidth]{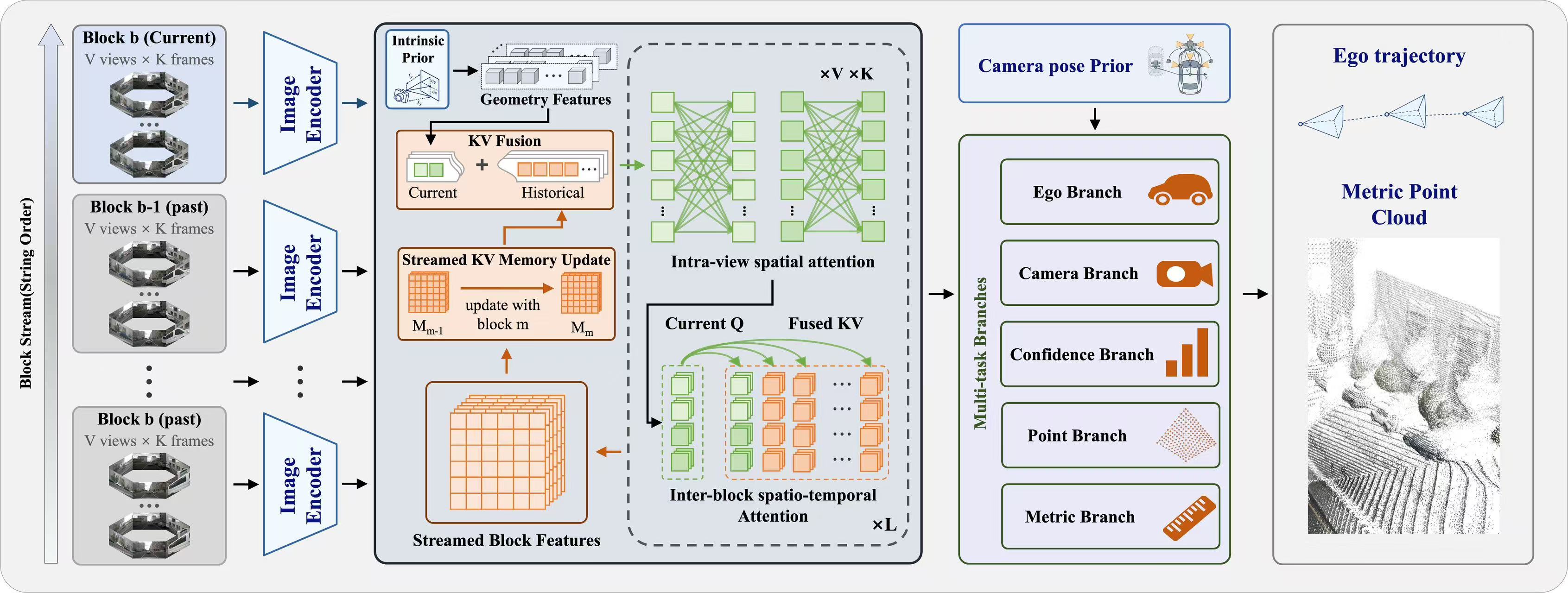} 
    
    \caption{\textbf{The pipeline of our method.} Multi-view image streams are processed in a block-wise manner. Each image block is first encoded with intrinsic priors and then processed by a spatio-temporal decoder. By maintaining a streamed KV memory across consecutive blocks, the model continuously extracts consistent geometry features for multi-task predictions, including ego-centric motion and dense metric 3D reconstruction.}
    \label{fig:overall_architecture}
\end{figure}

\noindent\textbf{Intrinsic-aware image encoding.}
As illustrated in Fig. \ref{fig:overall_architecture}, we adopt an encoder-decoder architecture for long-horizon surround-view image streams. For each image \(I_{t,v}\), a vision Transformer encoder extracts patch-level visual tokens. We further inject camera-intrinsic information by converting \(K_{t,v}\) into normalized camera-ray embeddings and adding them to the visual tokens. This provides explicit projection-geometry priors for dense 3D prediction.

\noindent\textbf{Spatio-temporal decoder.}
The intrinsic-aware tokens are then fed into a spatio-temporal Transformer decoder. The decoder adopts an alternating attention structure: intra-view spatial attention first models local context within each camera image, while inter-block spatio-temporal attention propagates information across all camera views and the temporally available blocks, jointly modeling multi-camera geometric consistency and temporal continuity. Unlike generic reconstruction models that apply fully bidirectional global attention to short input sequences, our decoder is designed for long driving logs and supports block-wise streaming inference. It preserves sufficient multi-frame and multi-camera interactions within each local temporal block, while propagating historical information along long sequences through streaming states. The block-wise streaming mechanism is introduced in Sec.~\ref{sec:block_streaming}.

\noindent\textbf{Task-specific prediction branches.}
The shared features produced by the spatio-temporal decoder are fed into several lightweight task-specific prediction branches, each adapting the shared representation to a particular geometric target. The dense point branch predicts per-view 3D point clouds \(\{P_{t,v}\}_{t=1,v=1}^{T,V}\), while a metric scale branch estimates a global scale factor that maps normalized geometry to real-world metric space. An auxiliary camera pose branch provides camera-pose supervision during training to improve geometric consistency. In addition, we introduce an ego-centric pose branch to estimate the vehicle-level trajectory \(\{E_t\}_{t=1}^{T}\). This branch uses the camera-to-ego extrinsics \(T^{\mathrm{ego}}_{t,v}\) as a rig-geometry prior during multi-camera feature aggregation, enabling observations from different cameras to be fused into a unified ego-centric motion representation. Finally, a block-level confidence branch predicts the reliability of the current streaming state and provides the trigger signal for confidence-aware re-anchoring.

\subsection{Block-wise Streaming Transformer}
\label{sec:block_streaming}

\noindent\textbf{Block-wise sequence partition.}
To extend a globally bidirectional geometric foundation model to long-horizon autonomous driving sequences, we partition the input stream into consecutive temporal blocks. Let each block contain \(K\) frames. The \(b\)-th block is defined as
\begin{equation}
    \mathcal{B}_b =
    \{(t,v)\mid t=(b-1)K+1,\ldots,\min(bK,T),\ v=1,\ldots,V\},
    \quad
    b=1,\ldots,\left\lceil \frac{T}{K} \right\rceil .
    \label{eq:block_partition}
\end{equation}
This partition preserves the ordered temporal structure and the regular surround-view camera layout of driving data. Unlike frame-wise causal modeling \cite{zhuo2025streaming}, each block contains multiple consecutive frames and all camera views, allowing the model to exploit local multi-frame and multi-camera context for geometric reasoning.

\noindent\textbf{Block-wise causal attention.}
During training, we process a sequence containing multiple temporal blocks in a single forward pass. To make this offline training procedure consistent with streaming inference, we introduce a block-wise causal mask in each inter-block spatio-temporal attention layer. For a query token \(q\) and a key token \(k\), let \(b(q)\) and \(b(k)\) denote their corresponding temporal block indices. The attention mask is defined as
\begin{equation}
M(q,k)=
\left\{
\begin{array}{ll}
0, & \text{if } b(k)\leq b(q),\\
-\infty, & \text{if } b(k)>b(q).
\end{array}
\right.
\label{eq:block_causal_mask}
\end{equation}
As a result, all tokens within the same block are mutually visible, including tokens from different timestamps and camera views, while tokens from future blocks are masked out. This training mask preserves bidirectional context within each block and enforces strict causality across blocks. Therefore, the model learns under the same information-access pattern that will be used during KV-cache streaming inference.

\noindent\textbf{Streaming inference with KV cache.}
During inference, the model processes the sequence block by block and maintains historical key/value caches in the inter-block  spatio-temporal attention
layers. For the \(b\)-th block, the current queries attend to both the key/value pairs from the current block and the cached key/value pairs from previous blocks. After processing the block, its key/value pairs are appended to the cache for future inference. Formally, for the \(\ell\)-th global attention layer, we have
\begin{equation}
    K_{\leq b}^{(\ell)}
    =
    \mathrm{Concat}\left(K_{<b}^{(\ell)}, K_b^{(\ell)}\right),\quad
    V_{\leq b}^{(\ell)}
    =
    \mathrm{Concat}\left(V_{<b}^{(\ell)}, V_b^{(\ell)}\right).
\end{equation}
By reusing historical features through the cache, the model avoids recomputing the entire history whenever a new block arrives. This enables efficient incremental streaming inference, while the model retains bidirectional reasoning within each block and causal streaming across blocks.

\subsection{Confidence-Aware Re-Anchoring}
\label{sec:confidence_reanchoring}

\noindent\textbf{Motivation.}
The block-wise streaming Transformer preserves bidirectional context within each local temporal block and propagates historical information over long sequences through KV cache. However, in long-horizon autonomous driving inference, cache propagation alone cannot completely prevent error accumulation. As the vehicle keeps moving and the scene content continuously changes, the implicit geometric state stored in the historical cache may gradually deviate from the current observations, eventually leading to ego-trajectory drift and point-cloud misalignment. To address this issue, we introduce a confidence-aware re-anchoring mechanism, enabling the model to online assess the reliability of the current streaming state and actively reset and realign when potential drift occurs.

\noindent\textbf{Block-level confidence prediction.}
To estimate the streaming reliability of each block, we introduce a block-level confidence branch on top of the shared spatio-temporal features. This branch first uses a lightweight decoder to extract geometry-state features related to the current block, then aggregates them over the spatial and camera dimensions, and outputs one confidence score \(c_b\) for each temporal block. A larger \(c_b\) indicates that the current block remains consistent with the existing streaming state, while a smaller \(c_b\) suggests that the current KV cache may have drifted. Unlike pixel-level confidence for dense geometry, the block-level confidence does not measure the reconstruction quality of individual points. Instead, it serves as a control signal for deciding whether the long-horizon streaming state is still reliable.

\noindent\textbf{Adaptive cache reset.}
During inference, the model processes the input sequence block by block and reads the confidence score after each block. If \(c_b\geq\tau\), where \(\tau\) is a predefined threshold, the model continues to reuse the current KV cache and appends the current block output to the global result. If \(c_b<\tau\), the model regards the current streaming state as potentially unreliable, discards the existing KV cache, and reinitializes a new local segment from overlapping frames near the current block. Unlike fixed-length chunk-then-align methods \cite{deng2025vggt, lin2025depth}, the restart position is adaptively determined by the model-predicted confidence, allowing the inference process to adjust according to scene difficulty and drift risk.

\noindent\textbf{SE(3) re-anchoring.}
To stitch the reinitialized local segment back to the existing global trajectory, we perform SE(3) alignment using the predicted poses on overlapping frames. Let \(\Omega_b\) denote the set of overlapping frames between the old and new segments. For each overlapping frame \(i\), we denote the predicted pose from the old trajectory as \((R_i^{\mathrm{old}}, t_i^{\mathrm{old}})\) and that from the new segment as \((R_i^{\mathrm{new}}, t_i^{\mathrm{new}})\). We estimate a rigid transform \(T_b^\star=(R_b^\star,t_b^\star)\in \mathrm{SE}(3)\) by
\begin{equation}
    R_b^\star =
    \Pi_{\mathrm{SO}(3)}
    \left(
    \frac{1}{|\Omega_b|}
    \sum_{i\in\Omega_b}
    R_i^{\mathrm{old}}
    \left(R_i^{\mathrm{new}}\right)^\top
    \right),
    \label{eq:se3_reanchoring_rotation}
\end{equation}
\begin{equation}
    t_b^\star =
    \frac{1}{|\Omega_b|}
    \sum_{i\in\Omega_b}
    \left(
    t_i^{\mathrm{old}} - R_b^\star t_i^{\mathrm{new}}
    \right),
    \label{eq:se3_reanchoring_translation}
\end{equation}
where \(\Pi_{\mathrm{SO}(3)}(\cdot)\) projects a matrix to the closest valid rotation. The resulting transform is then applied to the poses and point clouds of the new segment, re-anchoring them into the existing global coordinate frame.

\subsection{Training Losses}
\label{sec:training_losses}
Following $\pi^3\mathrm{x}$ \cite{wang2025pi}, we adopt standard geometric supervision for dense reconstruction, metric scale prediction, and auxiliary camera pose estimation. The dense geometry loss \(\mathcal{L}_{\mathrm{geo}}\) supervises predicted point maps and depths on valid pixels, while the metric scale loss \(\mathcal{L}_{\mathrm{scale}}\) encourages the predicted geometry to recover metric-space scale. We also include an auxiliary camera pose loss \(\mathcal{L}_{\mathrm{cam}}\) to improve the geometric consistency of the shared representation. In addition, the ego pose branch is supervised by \(\mathcal{L}_{\mathrm{ego}}\). Similar to camera pose supervision, \(\mathcal{L}_{\mathrm{ego}}\) combines translation error and geodesic rotation error for vehicle-level trajectory estimation.

For the block-level confidence branch, we derive the confidence target from trajectory drift errors. For the ego track, we take the first frame of the sequence segment as the reference frame and compute both predicted and ground-truth poses relative to this frame. We then measure their discrepancy using a weighted combination of translation error and rotation error, average it over frames within the \(b\)-th temporal block to obtain a block-level error \(e_b\). The soft confidence target is defined as
\begin{equation}
    y_b=\exp(-\alpha_{\mathrm{conf}} e_b),
\end{equation}
where \(\alpha_{\mathrm{conf}}\) is a fixed temperature coefficient, set to \(25\) in our experiments. Thus, a block with small relative trajectory error receives a confidence target close to 1, whereas a block with larger drift receives a lower confidence target. Given the predicted confidence logit \(s_b\), we use binary cross-entropy for supervision:
\begin{equation}
    \mathcal{L}_{\mathrm{conf}}
    =
    \mathrm{BCEWithLogits}(s_b,y_b).
\end{equation}

\section{Experiments}
\label{sec:experiments}

We evaluate the proposed block-wise streaming framework on Waymo \cite{sun2020scalability}, nuScenes \cite{caesar2020nuscenes}, and KITTI \cite{geiger_vision_2013}. Waymo \cite{sun2020scalability} and nuScenes \cite{caesar2020nuscenes} are used for dense geometry and ego-motion evaluation, while KITTI \cite{geiger_vision_2013} is used for long-sequence ego-motion estimation. We report depth, point-cloud, and trajectory metrics, including AbsRel, $\delta < 1.25$, Accuracy, Completeness, ATE, and AUC@30. We further conduct ablation and block-wise analyses to validate our design choices. Unless otherwise specified, all methods are evaluated under the same input protocol and metric computation pipeline.

\subsection{Dense Geometry and Ego-Motion Estimation}
\label{sec:waymo_nuscenes}

We first evaluate the proposed method on Waymo \cite{sun2020scalability} and nuScenes \cite{caesar2020nuscenes} under the 15-frame, full-view setting. By utilizing all available camera views, this setting jointly measures depth estimation, point-cloud reconstruction, and ego-motion estimation.

\begin{table}[!b]
  \centering
  \caption{Quantitative comparison on Waymo \cite{sun2020scalability} under the 15-frame, full-view setting. Best results are \textbf{bolded}, and second best are \underline{underlined}. * indicates models predicting relative scale that are aligned to recover the metric scale.}
  \label{tab:waymo}
  \vspace{-0.2em}
  \small
  \begin{tabular}{l cccccc}
    \toprule
    \multirow{2}{*}{Model} 
    & \multicolumn{2}{c}{Depth}  
    & \multicolumn{2}{c}{Point Cloud}  
    & \multicolumn{2}{c}{Ego-Motion} \\
    \cmidrule(lr){2-3} \cmidrule(lr){4-5} \cmidrule(lr){6-7}
    & AbsRel $\downarrow$  
    & $\delta < 1.25 \uparrow$  
    & Acc. $\downarrow$  
    & Comp. $\downarrow$  
    & AUC@30 $\uparrow$  
    & ATE $\downarrow$ \\
    \midrule
    \rowcolor[gray]{0.93}
    \multicolumn{7}{l}{\textit{Offline}} \\
    VGGT* \cite{wang2025vggt}       & 0.2386 & 0.7657 & 0.9033 & 1.4851 & \underline{86.78} & \textbf{3.2108} \\
    $\pi^3\mathrm{x}$ \cite{wang2025pi}        & 0.5005 & 0.3273 & \underline{0.7513} & 1.5304 & 77.04 & 6.9732 \\
    DA3-Nested \cite{lin2025depth} & 0.3970 & 0.3629 & 1.0020 & 2.0973 & 79.20 & \underline{3.3885} \\
    DVGT \cite{zuo2025dvgt}       & 0.4698 & 0.4183 & 0.9653 & 1.6218 & 85.58 & 3.6566 \\
    \midrule
    \rowcolor[gray]{0.93}
    \multicolumn{7}{l}{\textit{Per-frame Streaming}} \\
    StreamVGGT* \cite{zhuo2025streaming} & 0.3253 & 0.5616 & 1.0984 & 0.9806 & 82.72 & 4.9199 \\
    LongStream \cite{cheng_longstream:_2026}  & \underline{0.2065} & \underline{0.7849} & 0.9836 & 4.7566 & 40.88 & 9.7395 \\
    DVGT-2 \cite{zuo2026dvgt}      & 0.4356 & 0.4774 & 1.4894 & \underline{0.9539} & 84.73 & 3.8579 \\
    \midrule
    \rowcolor[gray]{0.93}
    \multicolumn{7}{l}{\textit{Block-wise Streaming}} \\
    \rowcolor[rgb]{0.90,0.96,1.00}
    \textbf{Ours}        & \textbf{0.1947} & \textbf{0.8429} & \textbf{0.5941} & \textbf{0.5894} & \textbf{86.84} & 3.9387 \\
    \bottomrule
  \end{tabular}

  \par\vspace{8pt}
  \centering
  \caption{Quantitative comparison on nuScenes \cite{caesar2020nuscenes} under the 15-frame, full-view setting.}
  \label{tab:nuscenes}
  \vspace{-0.2em}
  \small
  \begin{tabular}{l cccccc}
    \toprule
    \multirow{2}{*}{Model} 
    & \multicolumn{2}{c}{Depth}  
    & \multicolumn{2}{c}{Point Cloud}  
    & \multicolumn{2}{c}{Ego-Motion} \\
    \cmidrule(lr){2-3} \cmidrule(lr){4-5} \cmidrule(lr){6-7}
    & AbsRel $\downarrow$  
    & $\delta < 1.25 \uparrow$  
    & Acc. $\downarrow$  
    & Comp. $\downarrow$  
    & AUC@30 $\uparrow$  
    & ATE $\downarrow$ \\
    \midrule
    \rowcolor[gray]{0.93}
    \multicolumn{7}{l}{\textit{Offline}} \\
    VGGT* \cite{wang2025vggt}       & 0.1967 & 0.6835 & 0.8315 & 1.5516 & 85.47 & \underline{0.9151} \\
    $\pi^3\mathrm{x}$ \cite{wang2025pi}       & 0.2651 & 0.6672 & 0.9515 & 0.7946 & 78.88 & 3.1063 \\
    DA3-Nested \cite{lin2025depth} & 0.2261 & 0.6778 & 0.8521 & 0.9007 & 86.86 & 1.1709 \\
    DVGT \cite{zuo2025dvgt}        & \underline{0.1397} & \underline{0.8211} & \underline{0.6158} & \underline{0.6013} & \textbf{88.64} & 1.1012 \\
    \midrule
    \rowcolor[gray]{0.93}
    \multicolumn{7}{l}{\textit{Per-frame Streaming}} \\
    StreamVGGT* \cite{zhuo2025streaming} & 0.2623 & 0.4928 & 1.1090 & 2.4368 & 80.98 & 1.5748 \\
    LongStream \cite{cheng_longstream:_2026}  & 0.3026 & 0.4517 & 1.1155 & 0.9305 & 39.29 & 1.8030 \\
    DVGT-2 \cite{zuo2026dvgt}      & 0.1806 & 0.7285 & 0.6750 & 0.6230 & 85.08 & 1.5812 \\
    \midrule
    \rowcolor[gray]{0.93}
    \multicolumn{7}{l}{\textit{Block-wise Streaming}} \\
    \rowcolor[rgb]{0.90,0.96,1.00}
    \textbf{Ours}        & \textbf{0.1081} & \textbf{0.9077} & \textbf{0.5344} & \textbf{0.5374} & \underline{86.92} & \textbf{0.9092} \\
    \bottomrule
  \end{tabular}
  \vspace{-0.6em}
\end{table}

\noindent\textbf{Quantitative Evaluation.}
As shown in Tabs.~\ref{tab:waymo} and \ref{tab:nuscenes}, our method obtains the best depth and point-cloud reconstruction results on Waymo \cite{sun2020scalability}, with an AbsRel of 0.1947, $\delta < 1.25$ of 0.8429, Accuracy of 0.5941, and Completeness of 0.5894, while also achieving the highest AUC@30. On nuScenes \cite{caesar2020nuscenes}, it achieves the best results in five out of six metrics, including AbsRel, $\delta < 1.25$, Accuracy, Completeness, and ATE, showing consistent gains in both dense reconstruction and trajectory estimation.

\begin{figure}[!tbp]
    \centering
    \includegraphics[width=\textwidth]{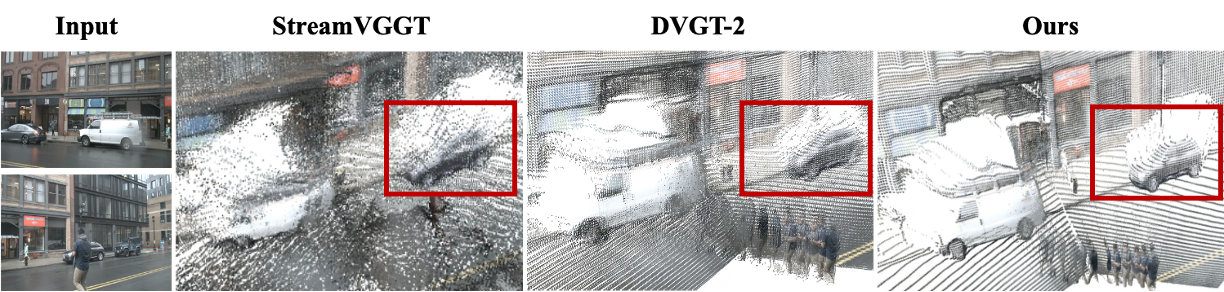}
    \caption{Qualitative comparison with per-frame streaming baselines. Red boxes highlight representative local regions where our method recovers cleaner geometry and sharper object structures.}
    \label{fig:qual_compare}
\end{figure}

\noindent\textbf{Qualitative Evaluation}
Fig.~\ref{fig:qual_compare} illustrates the geometric degradation in per-frame streaming baselines when bidirectional context is entirely discarded. Driven by the continuous accumulation of historical errors under a strict causal structure, StreamVGGT \cite{zhuo2025streaming} produces noisy and incomplete geometry. Similarly, while DVGT-2 \cite{zuo2026dvgt} preserves the coarse scene layout, it still blurs object boundaries and fine-grained structures. In contrast, by establishing bidirectional local contexts to halt temporal error amplification, our block-wise streaming method reconstructs cleaner road surfaces, sharper vehicle shapes, and more consistent foreground structures, demonstrating superior geometric coherence.

\subsection{Long-sequence Ego-Motion Estimation}
\label{sec:kitti_tracking}

We next evaluate long-sequence ego-motion estimation on KITTI \cite{geiger_vision_2013} using complete driving sequences. This setting stresses offline global-attention models, whose memory cost prevents single-pass inference over complete driving logs.
For KITTI \cite{geiger_vision_2013} stereo evaluation, we disable loop closure for methods whose loop module assumes a single-view temporal stream, since stereo inputs interleave synchronized left/right images rather than forming a purely monocular sequence. 
This setting therefore evaluates the long-range stability of each method under a consistent feed-forward odometry protocol. 
Tab.~\ref{tab:kitti} reports the ATE on all 11 KITTI \cite{geiger_vision_2013} sequences.
\begin{table}[!tbp]
  \centering
  \caption{Quantitative comparison on KITTI \cite{geiger_vision_2013} Stereo 2Hz. The metric is ATE after $\text{Sim}(3)$ alignment. $^\dagger$Average over completed sequences only; missing StreamVGGT \cite{zhuo2025streaming} entries failed with OOM.}
  \label{tab:kitti}
  \footnotesize
  \setlength{\tabcolsep}{2pt}
  \renewcommand{\arraystretch}{1.1}
  \resizebox{\textwidth}{!}{%
  \begin{tabular}{l *{11}{c} c}
    \toprule
    \multirow{3}{*}{\textbf{Methods}} 
    & \multicolumn{11}{c}{\textbf{KITTI \cite{geiger_vision_2013} Stereo 2Hz} (ATE$\downarrow$)} 
    & \multirow{3}{*}{\textbf{Avg. $\downarrow$}} \\
    \cmidrule(lr){2-12}
    & 00 & 01 & 02 & 03 & 04 & 05 & 06 & 07 & 08 & 09 & 10 & \\
    & \makecell[c]{$T{=}909$\\[-1pt]$V{=}2$\\[-1pt] 3.7km}
    & \makecell[c]{$T{=}221$\\[-1pt]$V{=}2$\\[-1pt] 2.5km}
    & \makecell[c]{$T{=}933$\\[-1pt]$V{=}2$\\[-1pt] 5.1km}
    & \makecell[c]{$T{=}161$\\[-1pt]$V{=}2$\\[-1pt] 0.6km}
    & \makecell[c]{$T{=}55$\\[-1pt]$V{=}2$\\[-1pt] 0.4km}
    & \makecell[c]{$T{=}553$\\[-1pt]$V{=}2$\\[-1pt] 2.2km}
    & \makecell[c]{$T{=}221$\\[-1pt]$V{=}2$\\[-1pt] 1.2km}
    & \makecell[c]{$T{=}221$\\[-1pt]$V{=}2$\\[-1pt] 0.7km}
    & \makecell[c]{$T{=}815$\\[-1pt]$V{=}2$\\[-1pt] 3.2km}
    & \makecell[c]{$T{=}319$\\[-1pt]$V{=}2$\\[-1pt] 1.7km}
    & \makecell[c]{$T{=}241$\\[-1pt]$V{=}2$\\[-1pt] 0.9km}
    & \\
    \midrule
    \rowcolor[gray]{0.93}
    \multicolumn{13}{l}{\textit{Per-frame Streaming}} \\
    CUT3R \cite{wang2025continuous}         & 181.0 & 692.4 & 271.6 & 51.3 & 16.2 & 147.5 & 133.3 & 84.8 & 256.1 & 155.6 & 63.0 & 186.6 \\
    TTT3R \cite{chen2025ttt3r}         & 181.1 &  68.6 & 243.2 & 16.2 &  6.3 & 138.8 & 127.5 & 64.9 & 126.3 & 110.4 & 43.3 & 102.4 \\
    DVGT-2 \cite{zuo2026dvgt}        & 148.4 & \underline{29.6} & 252.9 & 15.6 &  2.8 &  59.2 & \textbf{12.1} & \textbf{5.2} & 181.2 & 105.5 & 18.2 & 75.5 \\
    StreamVGGT \cite{zhuo2025streaming}     & OOM   & 680.6 & OOM   & 154.2 & 79.8 & OOM   & 123.4 & 79.3 & OOM   & 208.0 & 157.2 & 211.8$^\dagger$ \\
    \midrule
    \rowcolor[gray]{0.93}
    \multicolumn{13}{l}{\textit{chunk-then-Align}} \\
    DA3-Streaming \cite{lin2025depth,deng2025vggt} & 125.2 & 113.2 & 215.8 &  6.2 & \textbf{0.8} & 124.7 &  50.5 & 35.1 & 100.2 & 127.1 & 46.1 & 85.9 \\
    VGGT-Long \cite{deng2025vggt}     & 181.6 &  50.1 & 189.9 & \underline{4.9} & \underline{1.7} & \underline{43.3} & 33.3 & 21.8 & 96.8 & 120.9 & 21.0 & 69.6 \\
    $\pi$-Long \cite{wang2025pi,deng2025vggt}        & \textbf{71.7} & 167.1 & \textbf{129.2} &  7.5 &  2.4 & \textbf{33.2} & \underline{19.6} & \underline{7.3} & \textbf{61.4} & \underline{27.7} & \underline{15.0} & \underline{49.3} \\
    \midrule
    \rowcolor[gray]{0.93}
    \multicolumn{13}{l}{\textit{Block-wise Streaming}} \\
    \rowcolor[rgb]{0.90,0.96,1.00}
    \textbf{Ours}  & \underline{89.8} & \textbf{25.5} & \underline{135.8} & \textbf{4.6} & 2.5 & 52.4 & 44.2 & 7.8 & \underline{65.6} & \textbf{17.5} & \textbf{12.7} & \textbf{41.7} \\
    \bottomrule
  \end{tabular}%
  }
  \vspace{4pt}
  \raggedright
\end{table}

\begin{figure}[!tbp]
    \centering
    \includegraphics[width=\textwidth]{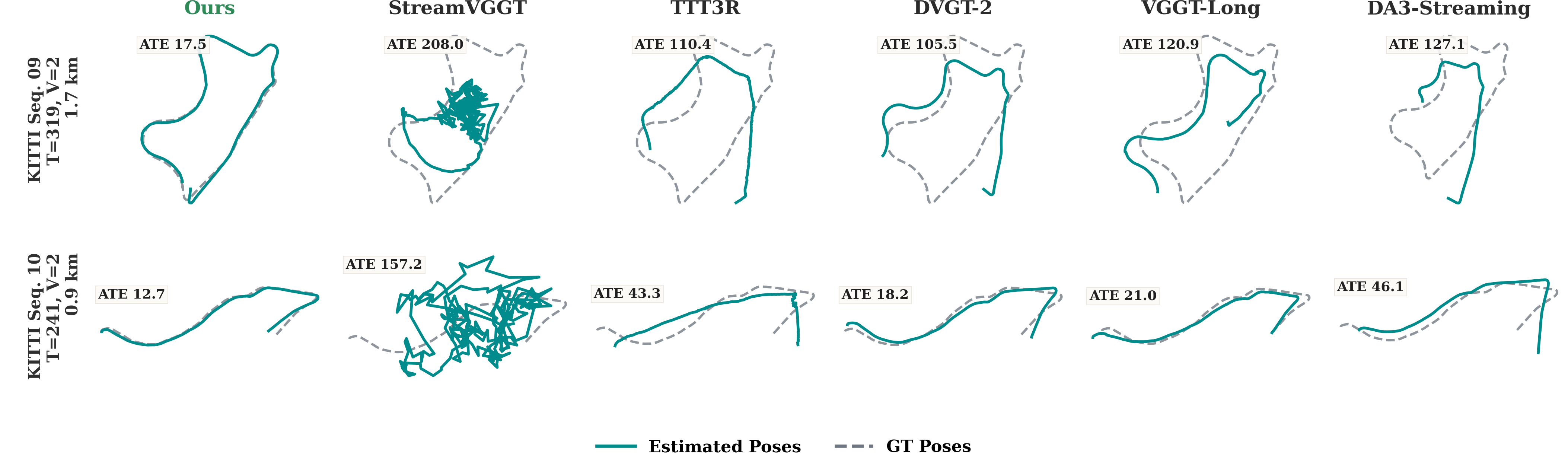}
    \caption{Qualitative trajectory comparison on KITTI \cite{geiger_vision_2013} sequences 09 and 10. The dashed gray curves denote ground-truth poses, and the solid teal curves denote estimated poses. Our method produces trajectories that are more consistent with the ground truth and obtains lower ATE.}
    \label{fig:kitti_traj}
\end{figure}

Our method achieves the best average ATE of 41.7, reducing the error by 44.8\% compared with DVGT-2 \cite{zuo2026dvgt}. The improvement suggests that block-wise streaming better preserves long-range temporal consistency than per-frame streaming and chunk-then-align baselines. This result also highlights the benefit of confidence-aware re-anchoring. Unlike StreamVGGT \cite{zhuo2025streaming}, whose cached history grows monotonically and can lead to OOM on long KITTI sequences, Drive3R detects unreliable streaming states and reinitializes the cache before stale history dominates the prediction. The refreshed segment is then stitched back through overlap alignment, yielding more stable long-range ego-motion estimation.  
The qualitative results in Fig.~\ref{fig:kitti_traj} further show that our estimated trajectories are more consistent with the ground truth on sequences 09 and 10.

\Needspace{14\baselineskip}
\subsection{Experimental Analysis}
\label{sec:analysis}

\setlength{\columnsep}{12pt}
\setlength{\intextsep}{4pt}
\begin{wraptable}{r}{0.54\textwidth}
\setlength{\tabcolsep}{2.2pt}
\renewcommand{\arraystretch}{1.10}
\renewcommand{\textbf}[1]{{\bfseries #1}}
\captionsetup{justification=raggedright,singlelinecheck=false}
\centering
\caption{Re-anchoring ablation on nuScenes \cite{caesar2020nuscenes}.}
\label{tab:ablation_confidence}
\fontsize{8.5}{10}\selectfont
\begin{tabular*}{\linewidth}{@{\extracolsep{\fill}}lcccccc@{}}
\toprule
\multirow{3}{*}{\textbf{Setting}}
& \multicolumn{2}{c}{\textbf{Depth}}
& \multicolumn{2}{c}{\textbf{Point Cloud}}
& \multicolumn{2}{c}{\textbf{Ego-Motion}} \\
\cmidrule(lr){2-3} \cmidrule(lr){4-5} \cmidrule(lr){6-7}
& AbsRel & $\delta < 1.25$ & Acc. & Comp. & AUC@30 & ATE \\[-2pt]
& $\downarrow$ & $\uparrow$ & $\downarrow$ & $\downarrow$ & $\uparrow$ & $\downarrow$ \\
\midrule
w/o Conf. & 0.1730 & 0.8274 & 0.8320 & 0.7918 & 70.66 & 18.1769 \\
\addlinespace[2pt]
\textbf{w/ Conf.} & \textbf{0.1063} & \textbf{0.9039} & \textbf{0.7129} & \textbf{0.6137} & \textbf{88.16} & \textbf{4.1422} \\
\bottomrule
\end{tabular*}
\end{wraptable}

\textbf{Ablation on confidence-aware re-anchoring.}
We first ablate the confidence-aware re-anchoring strategy on 36-frame, 6-view nuScenes \cite{caesar2020nuscenes} sequences. The efficacy of confidence-aware re-anchoring is demonstrated in Tab.~\ref{tab:ablation_confidence}. 
Removing this strategy triggers a significant performance drop, with ATE escalating from 4.1422 to 18.1769. 
This gap highlights the necessity of dynamic anchor selection. By identifying and re-anchoring to reliable temporal states, Drive3R effectively suppresses scale drift and trajectory deviation during streaming.

\Needspace{15\baselineskip}
\noindent\begin{minipage}[t]{0.48\linewidth}
\vspace{0pt}
\textbf{Analysis on block length.} We further analyze the influence of block length in Fig.~\ref{fig:block_metrics}. 
Here, CD denotes the sum of Accuracy and Completeness errors. 
Increasing the block size from 1 to 3 consistently improves Accuracy, Completeness, and CD, indicating that moderate temporal context benefits local geometry. 
Further increasing the block length improves Accuracy but hurts Completeness, suggesting a trade-off between local precision and scene coverage. 
Considering this trade-off, we choose a block size of 3 for our system. 
The OOM result of the offline setting further supports the necessity of bounded block-wise processing.

\end{minipage}\hfill
\begin{minipage}[t]{0.48\linewidth}
\vspace{0pt}
\centering
\begin{tikzpicture}
\begin{axis}[
    width=\linewidth,
    height=0.66\linewidth,
    xlabel={Frames per Block},
    ylabel={Metric},
    xmin=0,
    xmax=15,
    ymin=0.6,
    ymax=2.1,
    xtick={1,2,3,4,6,9,12,14},
    xticklabels={1,2,3,4,6,9,12,Offline},
    ymajorgrids=true,
    xmajorgrids=true,
    grid style={dashed, gray!30},
    legend style={
        at={(0.5, 1.04)},
        anchor=south,
        legend columns=3,
        font=\scriptsize,
        draw=none,
        fill=none,
        /tikz/every even column/.append style={column sep=0.15cm}
    },
    tick label style={font=\scriptsize},
    label style={font=\scriptsize},
    line width=1.0pt,
    mark size=2.0pt,
    clip=false,
]

\fill[gray!10] (axis cs:13, 0.6) rectangle (axis cs:15, 2.1);
\draw[dashed, gray!60, line width=0.8pt] (axis cs:13, 0.6) -- (axis cs:13, 2.1);

\addplot[
    blue!70!black,
    mark=*,
    mark options={fill=white, solid, line width=1pt}
] coordinates {
    (1,0.9084) (2,0.8843) (3,0.8320) (4,0.8302) (6,0.8088) (9,0.7803) (12,0.7564)
};
\addlegendentry{Acc.}

\addplot[
    orange!80!black,
    mark=square*,
    mark options={fill=white, solid, line width=1pt}
] coordinates {
    (1,1.0761) (2,0.9099) (3,0.7918) (4,0.7785) (6,0.8220) (9,0.9540) (12,0.9607)
};
\addlegendentry{Comp.}

\addplot[
    teal!80!black,
    mark=triangle*,
    mark options={fill=white, solid, line width=1pt}
] coordinates {
    (1,1.9845) (2,1.7942) (3,1.6238) (4,1.6087) (6,1.6308) (9,1.7343) (12,1.7171)
};
\addlegendentry{CD}

\node[
    font=\scriptsize\bfseries,
    rotate=90,
    anchor=center,
    text=red!70!black
] at (axis cs:14, 1.35) {OOM};

\end{axis}
\end{tikzpicture}

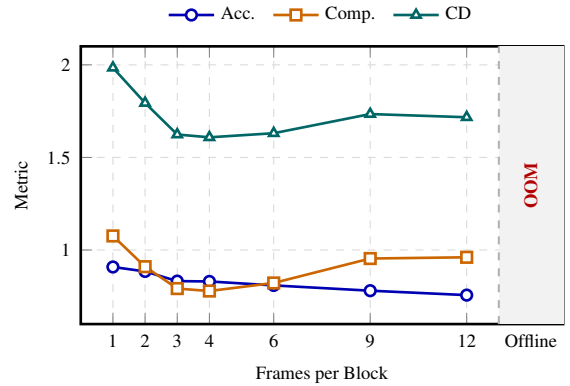
\captionof{figure}{Block-length analysis on nuScenes \cite{caesar2020nuscenes}.}
\label{fig:block_metrics}
\end{minipage}
\Needspace{4\baselineskip}
\section{Conclusion}

We present Drive3R, a block-wise streaming geometric foundation model for long-horizon autonomous driving reconstruction. Drive3R bridges offline bidirectional reconstruction and causal streaming by preserving local multi-frame and multi-camera reasoning within each block while reusing historical features through a KV cache, with confidence-aware re-anchoring to correct residual long-range drift. Extensive experiments show strong performance on dense depth, point-cloud reconstruction, and ego-motion estimation, demonstrating the value of adapting feed-forward geometric foundation models to continuous driving logs. Looking forward, Drive3R can serve as a foundation for online Gaussian scene representations and closed-loop autonomous driving simulation.

\FloatBarrier
\bibliographystyle{plainnat}
\bibliography{main}

\begin{thebibliography}{34}
\providecommand{\natexlab}[1]{#1}
\providecommand{\url}[1]{\texttt{#1}}
\expandafter\ifx\csname urlstyle\endcsname\relax
  \providecommand{\doi}[1]{doi: #1}\else
  \providecommand{\doi}{doi: \begingroup \urlstyle{rm}\Url}\fi

\bibitem[Caesar et~al.(2020)Caesar, Bankiti, Lang, Vora, Liong, Xu, Krishnan, Pan, Baldan, and Beijbom]{caesar2020nuscenes}
Holger Caesar, Varun Bankiti, Alex~H Lang, Sourabh Vora, Venice~Erin Liong, Qiang Xu, Anush Krishnan, Yu~Pan, Giancarlo Baldan, and Oscar Beijbom.
\newblock {nuScenes}: A multimodal dataset for autonomous driving.
\newblock In \emph{Proceedings of the IEEE/CVF conference on computer vision and pattern recognition}, pages 11621--11631, 2020.

\bibitem[Chen et~al.(2025)Chen, Chen, Xiu, Geiger, and Chen]{chen2025ttt3r}
Xingyu Chen, Yue Chen, Yuliang Xiu, Andreas Geiger, and Anpei Chen.
\newblock Ttt3r: 3d reconstruction as test-time training.
\newblock \emph{arXiv preprint arXiv:2509.26645}, 2025.

\bibitem[Cheng et~al.(2026)Cheng, Chen, Xie, Yin, Ren, Zhang, Guo, and Wang]{cheng_longstream:_2026}
Chong Cheng, Xianda Chen, Tao Xie, Wei Yin, Weiqiang Ren, Qian Zhang, Xiaoyuang Guo, and Hao Wang.
\newblock {LongStream}: {Long}-{Sequence} {Streaming} {Autoregressive} {Visual} {Geometry}.
\newblock \emph{arXiv preprint arXiv:2602.13172}, February 2026.

\bibitem[Deng et~al.(2025)Deng, Ti, Xu, Yang, and Xie]{deng2025vggt}
Kai Deng, Zexin Ti, Jiawei Xu, Jian Yang, and Jin Xie.
\newblock Vggt-long: Chunk it, loop it, align it--pushing vggt's limits on kilometer-scale long rgb sequences.
\newblock \emph{arXiv preprint arXiv:2507.16443}, 2025.

\bibitem[Geiger et~al.(2013)Geiger, Lenz, Stiller, and Urtasun]{geiger_vision_2013}
A~Geiger, P~Lenz, C~Stiller, and R~Urtasun.
\newblock Vision meets robotics: {The} {KITTI} dataset.
\newblock \emph{The International Journal of Robotics Research}, 32\penalty0 (11):\penalty0 1231--1237, September 2013.
\newblock \doi{10.1177/0278364913491297}.

\bibitem[Guo et~al.(2025)Guo, Zhang, He, Wang, Jing, and Ling]{guo2025end}
Mingzhe Guo, Zhipeng Zhang, Yuan He, Ke~Wang, Liping Jing, and Haibin Ling.
\newblock End-to-end autonomous driving without costly modularization and 3d manual annotation.
\newblock \emph{IEEE Transactions on Pattern Analysis and Machine Intelligence}, 2025.

\bibitem[Hong et~al.(2023)Hong, Zhang, Gu, Bi, Zhou, Liu, Liu, Sunkavalli, Bui, and Tan]{hong2023lrm}
Yicong Hong, Kai Zhang, Jiuxiang Gu, Sai Bi, Yang Zhou, Difan Liu, Feng Liu, Kalyan Sunkavalli, Trung Bui, and Hao Tan.
\newblock Lrm: Large reconstruction model for single image to 3d.
\newblock \emph{arXiv preprint arXiv:2311.04400}, 2023.

\bibitem[Hu et~al.(2023)Hu, Yang, Chen, Li, Sima, Zhu, Chai, Du, Lin, Wang, et~al.]{hu2023planning}
Yihan Hu, Jiazhi Yang, Li~Chen, Keyu Li, Chonghao Sima, Xizhou Zhu, Siqi Chai, Senyao Du, Tianwei Lin, Wenhai Wang, et~al.
\newblock Planning-oriented autonomous driving.
\newblock In \emph{Proceedings of the IEEE/CVF conference on computer vision and pattern recognition}, pages 17853--17862, 2023.

\bibitem[Kerbl et~al.(2023)Kerbl, Kopanas, Leimk{\"u}hler, Drettakis, et~al.]{kerbl20233d}
Bernhard Kerbl, Georgios Kopanas, Thomas Leimk{\"u}hler, George Drettakis, et~al.
\newblock 3d gaussian splatting for real-time radiance field rendering.
\newblock \emph{ACM Trans. Graph.}, 42\penalty0 (4):\penalty0 139--1, 2023.

\bibitem[Lin et~al.(2025)Lin, Chen, Liew, Chen, Li, Shi, Feng, and Kang]{lin2025depth}
Haotong Lin, Sili Chen, Junhao Liew, Donny~Y Chen, Zhenyu Li, Guang Shi, Jiashi Feng, and Bingyi Kang.
\newblock Depth anything 3: Recovering the visual space from any views.
\newblock \emph{arXiv preprint arXiv:2511.10647}, 2025.

\bibitem[Mildenhall et~al.(2021)Mildenhall, Srinivasan, Tancik, Barron, Ramamoorthi, and Ng]{mildenhall2021nerf}
Ben Mildenhall, Pratul~P Srinivasan, Matthew Tancik, Jonathan~T Barron, Ravi Ramamoorthi, and Ren Ng.
\newblock Nerf: Representing scenes as neural radiance fields for view synthesis.
\newblock \emph{Communications of the ACM}, 65\penalty0 (1):\penalty0 99--106, 2021.

\bibitem[Schonberger and Frahm(2016)]{schonberger2016structure}
Johannes~L Schonberger and Jan-Michael Frahm.
\newblock Structure-from-motion revisited.
\newblock In \emph{Proceedings of the IEEE conference on computer vision and pattern recognition}, pages 4104--4113, 2016.

\bibitem[Shen et~al.(2025)Shen, Zhang, Qu, Zheng, Ji, Zhang, and Cao]{shen2025fastvggt}
You Shen, Zhipeng Zhang, Yansong Qu, Xiawu Zheng, Jiayi Ji, Shengchuan Zhang, and Liujuan Cao.
\newblock Fastvggt: Training-free acceleration of visual geometry transformer.
\newblock \emph{arXiv preprint arXiv:2509.02560}, 2025.

\bibitem[Shi et~al.(2025)Shi, Shi, Lyu, Liu, Sheng, Zhang, and Jiang]{shi2025unisplat}
Chen Shi, Shaoshuai Shi, Xiaoyang Lyu, Chunyang Liu, Kehua Sheng, Bo~Zhang, and Li~Jiang.
\newblock Unisplat: Unified spatio-temporal fusion via 3d latent scaffolds for dynamic driving scene reconstruction.
\newblock \emph{arXiv preprint arXiv:2511.04595}, 2025.

\bibitem[Sun et~al.(2020)Sun, Kretzschmar, Dotiwalla, Chouard, Patnaik, Tsui, Guo, Zhou, Chai, Caine, et~al.]{sun2020scalability}
Pei Sun, Henrik Kretzschmar, Xerxes Dotiwalla, Aurelien Chouard, Vijaysai Patnaik, Paul Tsui, James Guo, Yin Zhou, Yuning Chai, Benjamin Caine, et~al.
\newblock Scalability in perception for autonomous driving: Waymo open dataset.
\newblock In \emph{2020 IEEE/CVF Conference on Computer Vision and Pattern Recognition (CVPR)}, pages 2443--2451. IEEE, 2020.

\bibitem[Tang et~al.(2025)Tang, Fan, Wang, Xu, Ranjan, Schwing, and Yan]{tang2025mv}
Zhenggang Tang, Yuchen Fan, Dilin Wang, Hongyu Xu, Rakesh Ranjan, Alexander Schwing, and Zhicheng Yan.
\newblock Mv-dust3r+: Single-stage scene reconstruction from sparse views in 2 seconds.
\newblock In \emph{Proceedings of the Computer Vision and Pattern Recognition Conference}, pages 5283--5293, 2025.

\bibitem[Tian et~al.(2023)Tian, Jiang, Yun, Mao, Yang, Wang, Wang, and Zhao]{tian2023occ3d}
Xiaoyu Tian, Tao Jiang, Longfei Yun, Yucheng Mao, Huitong Yang, Yue Wang, Yilun Wang, and Hang Zhao.
\newblock Occ3d: A large-scale 3d occupancy prediction benchmark for autonomous driving.
\newblock \emph{Advances in Neural Information Processing Systems}, 36:\penalty0 64318--64330, 2023.

\bibitem[Tonderski et~al.(2024)Tonderski, Lindstr{\"o}m, Hess, Ljungbergh, Svensson, and Petersson]{tonderski2024neurad}
Adam Tonderski, Carl Lindstr{\"o}m, Georg Hess, William Ljungbergh, Lennart Svensson, and Christoffer Petersson.
\newblock Neurad: Neural rendering for autonomous driving.
\newblock In \emph{Proceedings of the IEEE/CVF Conference on Computer Vision and Pattern Recognition}, pages 14895--14904, 2024.

\bibitem[Wang et~al.(2025{\natexlab{a}})Wang, Chen, Karaev, Vedaldi, Rupprecht, and Novotny]{wang2025vggt}
Jianyuan Wang, Minghao Chen, Nikita Karaev, Andrea Vedaldi, Christian Rupprecht, and David Novotny.
\newblock Vggt: Visual geometry grounded transformer.
\newblock In \emph{Proceedings of the Computer Vision and Pattern Recognition Conference}, pages 5294--5306, 2025{\natexlab{a}}.

\bibitem[Wang et~al.(2025{\natexlab{b}})Wang, Zhang, Holynski, Efros, and Kanazawa]{wang2025continuous}
Qianqian Wang, Yifei Zhang, Aleksander Holynski, Alexei~A Efros, and Angjoo Kanazawa.
\newblock Continuous 3d perception model with persistent state.
\newblock In \emph{Proceedings of the Computer Vision and Pattern Recognition Conference}, pages 10510--10522, 2025{\natexlab{b}}.

\bibitem[Wang et~al.(2024)Wang, Leroy, Cabon, Chidlovskii, and Revaud]{wang2024dust3r}
Shuzhe Wang, Vincent Leroy, Yohann Cabon, Boris Chidlovskii, and Jerome Revaud.
\newblock Dust3r: Geometric 3d vision made easy.
\newblock In \emph{Proceedings of the IEEE/CVF conference on computer vision and pattern recognition}, pages 20697--20709, 2024.

\bibitem[Wang et~al.(2025{\natexlab{c}})Wang, Zhou, Zhu, Chang, Zhou, Li, Chen, Pang, Shen, and He]{wang2025pi}
Yifan Wang, Jianjun Zhou, Haoyi Zhu, Wenzheng Chang, Yang Zhou, Zizun Li, Junyi Chen, Jiangmiao Pang, Chunhua Shen, and Tong He.
\newblock {{$\pi^3$}: Permutation-Equivariant Visual Geometry Learning}.
\newblock \emph{arXiv preprint arXiv:2507.13347}, 2025{\natexlab{c}}.

\bibitem[Yan et~al.(2024)Yan, Lin, Zhou, Wang, Sun, Zhan, Lang, Zhou, and Peng]{yan2024street}
Yunzhi Yan, Haotong Lin, Chenxu Zhou, Weijie Wang, Haiyang Sun, Kun Zhan, Xianpeng Lang, Xiaowei Zhou, and Sida Peng.
\newblock Street gaussians: Modeling dynamic urban scenes with gaussian splatting.
\newblock In \emph{European Conference on Computer Vision}, pages 156--173. Springer, 2024.

\bibitem[Yao et~al.(2018)Yao, Luo, Li, Fang, and Quan]{yao2018mvsnet}
Yao Yao, Zixin Luo, Shiwei Li, Tian Fang, and Long Quan.
\newblock Mvsnet: Depth inference for unstructured multi-view stereo.
\newblock In \emph{Proceedings of the European conference on computer vision (ECCV)}, pages 767--783, 2018.

\bibitem[Yu et~al.(2026)Yu, Xiao, Wang, Hao, Huang, Hu, Qin, Jing, Bo, and Luo]{yu2026recondrive}
Haibao Yu, Kuntao Xiao, Jiahang Wang, Ruiyang Hao, Yuxin Huang, Guoran Hu, Haifang Qin, Bowen Jing, Yuntian Bo, and Ping Luo.
\newblock Recondrive: Fast feed-forward 4d gaussian splatting for autonomous driving scene reconstruction.
\newblock \emph{arXiv preprint arXiv:2603.07552}, 2026.

\bibitem[Yuan et~al.(2026)Yuan, Yang, Yang, Zhang, Zhao, Zhang, and Zhang]{yuan2026infinitevggt}
Shuai Yuan, Yantai Yang, Xiaotian Yang, Xupeng Zhang, Zhonghao Zhao, Lingming Zhang, and Zhipeng Zhang.
\newblock Infinitevggt: Visual geometry grounded transformer for endless streams.
\newblock \emph{arXiv preprint arXiv:2601.02281}, 2026.

\bibitem[Zhang et~al.(2026{\natexlab{a}})Zhang, Gagvani, Cui, Peng, Zhang, and Wang]{zhang2026efficient}
Jiaru Zhang, Manav Gagvani, Can Cui, Juntong Peng, Ruqi Zhang, and Ziran Wang.
\newblock Efficient and explainable end-to-end autonomous driving via masked vision-language-action diffusion.
\newblock \emph{arXiv preprint arXiv:2602.20577}, 2026{\natexlab{a}}.

\bibitem[Zhang et~al.(2025)Zhang, Wu, Li, Ma, Ma, Li, and Wang]{zhang2025review}
Wei Zhang, Yihang Wu, Songhua Li, Wenjie Ma, Xin Ma, Qiang Li, and Qi~Wang.
\newblock Review of feed-forward 3d reconstruction: From dust3r to vggt.
\newblock \emph{arXiv preprint arXiv:2507.08448}, 2025.

\bibitem[Zhang et~al.(2026{\natexlab{b}})Zhang, Chen, Gao, Wang, Ge, Hu, Shi, and Zhang]{zhang2026onedrive}
Yiwei Zhang, Xuesong Chen, Jin Gao, Hanshi Wang, Fudong Ge, Weiming Hu, Shaoshuai Shi, and Zhipeng Zhang.
\newblock Onedrive: Unified multi-paradigm driving with vision-language-action models.
\newblock \emph{arXiv preprint arXiv:2604.17915}, 2026{\natexlab{b}}.

\bibitem[Zhou et~al.(2024)Zhou, Lin, Shan, Wang, Sun, and Yang]{zhou2024drivinggaussian}
Xiaoyu Zhou, Zhiwei Lin, Xiaojun Shan, Yongtao Wang, Deqing Sun, and Ming-Hsuan Yang.
\newblock Drivinggaussian: Composite gaussian splatting for surrounding dynamic autonomous driving scenes.
\newblock In \emph{Proceedings of the IEEE/CVF conference on computer vision and pattern recognition}, pages 21634--21643, 2024.

\bibitem[Zhou et~al.(2025)Zhou, Cai, Zhao, Zhang, Huang, Zhou, and Ma]{zhou2025autovla}
Zewei Zhou, Tianhui Cai, Seth~Z Zhao, Yun Zhang, Zhiyu Huang, Bolei Zhou, and Jiaqi Ma.
\newblock Autovla: A vision-language-action model for end-to-end autonomous driving with adaptive reasoning and reinforcement fine-tuning.
\newblock \emph{arXiv preprint arXiv:2506.13757}, 2025.

\bibitem[Zhuo et~al.(2025)Zhuo, Zheng, Guo, Wu, Zhou, and Lu]{zhuo2025streaming}
Dong Zhuo, Wenzhao Zheng, Jiahe Guo, Yuqi Wu, Jie Zhou, and Jiwen Lu.
\newblock Streaming 4d visual geometry transformer.
\newblock \emph{arXiv preprint arXiv:2507.11539}, 2025.

\bibitem[Zuo et~al.(2025)Zuo, Xie, Zheng, Xu, Li, Jiang, Chen, Yang, and Lu]{zuo2025dvgt}
Sicheng Zuo, Zixun Xie, Wenzhao Zheng, Shaoqing Xu, Fang Li, Shengyin Jiang, Long Chen, Zhi-Xin Yang, and Jiwen Lu.
\newblock Dvgt: Driving visual geometry transformer.
\newblock \emph{arXiv preprint arXiv:2512.16919}, 2025.

\bibitem[Zuo et~al.(2026)Zuo, Xie, Zheng, Xu, Li, Li, Chen, Yang, and Lu]{zuo2026dvgt}
Sicheng Zuo, Zixun Xie, Wenzhao Zheng, Shaoqing Xu, Fang Li, Hanbing Li, Long Chen, Zhi-Xin Yang, and Jiwen Lu.
\newblock Dvgt-2: Vision-geometry-action model for autonomous driving at scale.
\newblock \emph{arXiv preprint arXiv:2604.00813}, 2026.

\end{thebibliography}
\end{document}